%% file: Lee.tex
\documentclass[letterpaper]{article} % DO NOT CHANGE THIS
\usepackage{aaai24}  % DO NOT CHANGE THIS
\usepackage{times}  % DO NOT CHANGE THIS
\usepackage{helvet}  % DO NOT CHANGE THIS
\usepackage{courier}  % DO NOT CHANGE THIS
\usepackage[hyphens]{url}  % DO NOT CHANGE THIS
\usepackage{graphicx} % DO NOT CHANGE THIS
\usepackage{natbib}  % DO NOT CHANGE THIS AND DO NOT ADD ANY OPTIONS TO IT
\usepackage{caption} % DO NOT CHANGE THIS AND DO NOT ADD ANY OPTIONS TO IT
\usepackage{algorithm}
\usepackage{algorithmic}

\usepackage{newfloat}
\usepackage{listings}
\DeclareCaptionStyle{ruled}{labelfont=normalfont,labelsep=colon,strut=off} % DO NOT CHANGE THIS
\floatstyle{ruled}
\newfloat{listing}{tb}{lst}{}
\floatname{listing}{Listing}
\usepackage{graphicx}
\usepackage{amsmath}
\usepackage{amssymb}
\usepackage{multirow}
\usepackage{multicol}
\usepackage{cancel}
\usepackage{booktabs}

\usepackage[symbol]{footmisc}

\title{Noise-free Optimization in Early Training Steps for Image Super-Resolution}
\author{
    MinKyu Lee,
    Jae-Pil Heo\thanks{Corresponding author}
}
\affiliations{
    Sungkyunkwan University\\
    \{bluelati98, jaepilheo\}@g.skku.edu
}

\usepackage{bibentry}
\usepackage{ragged2e}

\begin{document}
\maketitle

\begin{abstract}
%
%%%%%%%%%%%%% (1) ----- Everybody uses this training scheme
% Recent deep-learning-based methods have shown impressive performance in low-level vision tasks by simply minimizing the pixel-wise distance with respect to a given high-resolution (HR) image as the unique ground truth.
Recent deep-learning-based single image super-resolution (SISR) methods have shown impressive performance whereas typical methods train their networks by minimizing the pixel-wise distance with respect to a given high-resolution (HR) image.
%
%%%%%%%%%%%%% (2) ----- But it is not well discovered
However, despite the basic training scheme being the predominant choice, its use in the context of ill-posed inverse problems has not been thoroughly investigated.
%
%%%%%%%%%%%%% (3) ----- We found that this leads to noisy training (especially in early training_)
In this work, we aim to provide a better comprehension of the underlying constituent by decomposing target HR images into two subcomponents: (1) the \textbf{optimal centroid} which is the expectation over multiple potential HR images, and (2) the \textbf{inherent noise} defined as the residual between the HR image and the centroid.
Our findings show that the current training scheme cannot capture the ill-posed nature of SISR and becomes vulnerable to the inherent noise term, especially during early training steps. 
%
%%%%%%%%%%%%% (4) ----- We solve this by optimizing towards to centroid.
To tackle this issue, we propose a novel optimization method that can effectively remove the inherent noise term in the early steps of vanilla training by estimating the optimal centroid and directly optimizing toward the estimation.
% 
%
%%%%%%%%%%%%% (5) ---- We are better
Experimental results show that the proposed method can effectively enhance the stability of vanilla training, leading to overall performance gain. Codes are available at github.com/2minkyulee/ECO.

\end{abstract}

\begin{figure*}[t]
    \begin{center}
    \includegraphics[width=\textwidth]{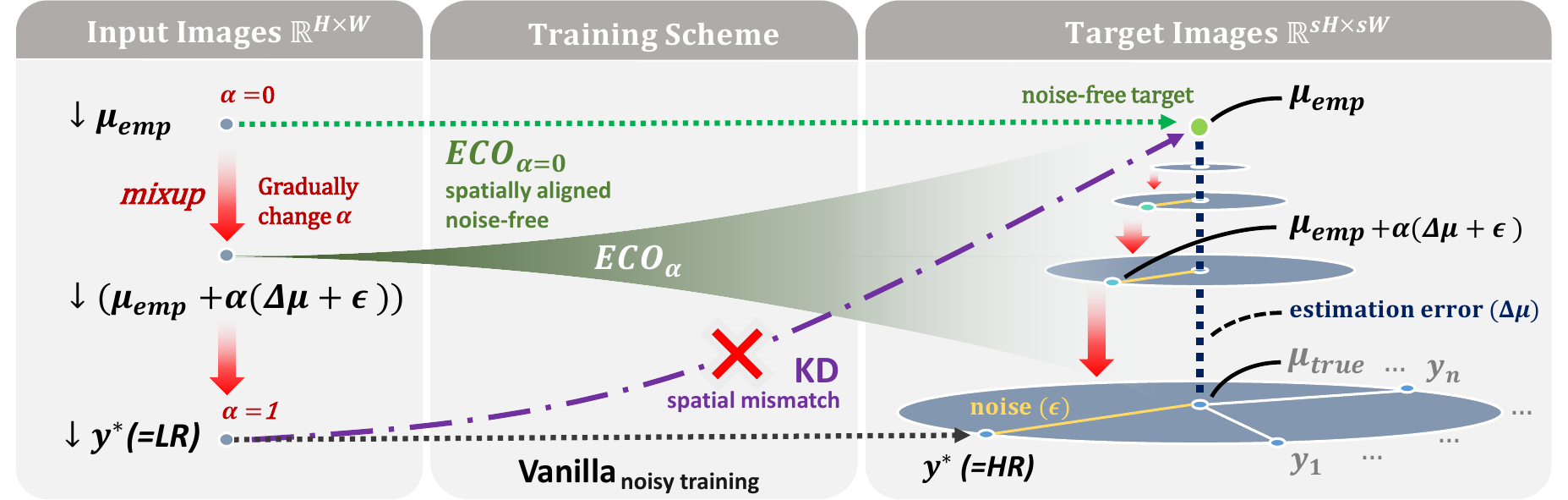}
    \end{center}
    \caption{
    Visualization of our method (ECO) compared to vanilla training and knowledge distillation (KD). Data points indicated in gray text are not available during training.
    Vanilla training leads to noisy training since it is unaware of the inherent noise $\epsilon$, which is defined as the difference of a given HR image $y^*$ and the expectation over all possible HR images, $\mu_{\text{true}}$. On the other hand, KD benefits from noise-free targets but suffers from spatial inconsistency between the input and target images as in Eq.\eqref{eq:construction_of_kd}.
    The proposed objective Eq.\eqref{eq:train_with_sr} benefits from noise-free training while being spatially aligned. Then, we overcome the limitations that arise by removing the estimation error term $\Delta\mu:=\mu_\text{true}-\mu_\text{emp}$ with a smooth transition from the proposed objective to the original objective. Remarkably, the overall solution can be greatly simplified with the use of \textit{mixup} strategy as in Eq.\eqref{eq:final_w_mixup} (Section \ref{section:mixup_as_rescue}). 
    Starting from synthetic data pairs ($\alpha=0$), gradually migrate to real data pairs ($\alpha=1$). This way, we enjoy noise-free training during the early steps, and finetune the network with supervision from real data samples in later steps.
    }   
    \label{fig:main_figure}
\end{figure*}

\section{Introduction}
With the drastic development of deep-learning-based techniques, recent single image super-resolution (SISR) methods have shown promising performance against previous methods. Here, the two primary objectives of SISR are; achieving precise reconstruction at the pixel level (known as fidelity-oriented methods); and producing visually appealing \cite{niqe, lpips} images (referred to as perceptual-quality-oriented methods). While perceptual-quality-oriented methods have become increasingly popular in recent years, fidelity-oriented methods still remain a mainstream of research due to the high demand for reliable reconstruction. Accordingly, we limit our focus to fidelity-oriented methods in this paper.

Typically, modern fidelity-oriented SISR networks adopt a very simple training strategy. In most cases, the only objective is to optimize the likelihood of the predicted image based on pairs of HR images and corresponding downscaled LR images. Here, with fair assumptions on the distribution of image spaces and empirical results \cite{SISR2_EDSR}, the majority decision of the objective function is narrowed down as the pixel-wise $L_1$ loss. However, although this basic training scheme is the predominant choice, its use and limitations have not been thoroughly investigated, particularly with regard to the ill-posed nature of image super-resolution. 

%%%%%%%%%%%%%%%%%%%%%%%%%%%%%%%%%%%%%%%%%%%%%%%%%%%%%%%%%% Our Contribution %%%%%%%%%%%%%%%%%%%%%%%%%%%%%%%%%%%%%%%%%%%%%%%%%%%%%%%%%%%%%%%%%%%%%
%%%%%%%%%%%%% (1) ----- We define optimal centroid and inherent noise
In this paper, we aim to analyze the underlying components of vanilla training in the context of SISR tasks and systematically develop the current training process. We start our analysis by decomposing the original HR image into two key components: \textit{optimal centroid} and \textit{inherent noise}. Given the ill-posed nature of image SR  \cite{hyun2020varsr, SISR9_SRFlow}, we define the optimal centroid as the expectation over multiple potential HR images that downsamples to an identical LR image instance. Additionally, we define the inherent noise term as the residual between the HR image sample and the optimal centroid, which is a fundamental component underlying in each HR image instance.

%%%%%%%%%%%%% (2) ----- We find that the noise term can hurt the training procedure
Our findings are that vanilla training neglects the ill-posed nature of inverse problems, which results as a residual noise term per sample.
Consequently, the overall training procedure becomes highly dependent on each HR image sample within a mini-batch, leading to noisy and unstable training, especially in early training steps.
% This urges us to remove the noise term for healthy and stable training.

%%%%%%%%%%%%% (3) ----- We solve this issue like this
In order to tackle this issue, we take the ill-posed nature of SR into account and formulate a noise-free objective, which simplifies as minimizing the $L_1$ distance between the network's estimation and the expectation over all possible HR samples (i.e., the true centroid term). However, since direct usage of this objective is impossible due to the intractability of the centroid term, we utilize a surrogate objective that can effectively act as a substitute for the intractable objective. Specifically, we estimate the true centroid by an empirical centroid obtained from pretrained SR networks and define a tractable objective for noise-free optimization. Further, we show that Knowledge Distillation (KD) can be understood as a specific case of this noise-free optimization, but with apparent flaws: spatial inconsistency. We make a quick fix for the shortcomings of KD and construct a noise-free training objective that optimizes directly towards the empirical centroid while being both tractable and spatially aligned. It turns out that the proposed objective can lead to well-behaving loss values and gradients (i.e., better Lipschitzness) enabling stable optimization which is especially beneficial in the early steps of training.  At last, we address the limitations that come from the estimation error and provide a simple method to overcome this. With a smooth transition between the proposed noise-free objective and the original loss, it is shown that the proposed training framework can benefit from stable training during early steps and minimize shortcomings of approximation errors in later steps.
 
%%%%%%%%%%%%% (4) ----- Summary and Ending
To sum up, the major contribution of this work is in offering improved comprehension of the underlying processes involved in training neural networks for SISR tasks. This is further extended to a novel training framework, which we refer to as \textbf{Empirical Centroid-oriented Optimization (ECO)}. Experimental results show that ECO can lead to performance gain against vanilla training by enabling stable training and providing well-behaving optimization landscapes, especially helpful in the early training stages.

\section{Probabilistic Modeling}
\subsection{Traditional objective function}
With plausible assumptions of the HR image manifold, the widely used MLE strategy in low-level vision tasks are formulated as minimizing the $L_p$-norm.
Here, the majority choice in SR tasks is the $L_1$-norm since it has been empirically shown to have better convergence than the $L_2$-norm \cite{SISR2_EDSR}.
Accordingly, typical methods employ pixel-wise $L_1$ loss as the objective function where each HR image sample from the training dataset is treated as the sole ground truth image.
Thus, it is a clear choice to construct the objective function in the form of a loss for a single data point as follows:
\begin{equation}
    \begin{aligned}
        &f:=R^{H \times W} \rightarrow R^{sH \times sW} \\
        &L_1(HR, SR) = ||y^* - f(x)||_1,
    \end{aligned}
\end{equation}
where $f(\cdot)$ is the SR network with scale-factor $s$, which is piece-wise linear \cite{PiecewiseLinear} with only ReLU \cite{relu} as the non-linearity, and $y^*$, $x$ each corresponds to the HR, LR image sample in the training dataset, respectively.

% --------------------------------------------------------------------- %
% --------------------------------------------------------------------- %
% --------------------------------------------------------------------- %

\subsection{Optimal centroid and inherent noise}
Before we start our analysis, we define two fundamental components of ill-posed inverse problems: (1) the optimal centroid $\mu_{\text{true}}$ which is the expectation over multiple plausible solutions, and (2) the inherent noise $\epsilon$ which is the residual between the optimal centroid and a single data point. Here, the inherent noise term $\epsilon$ can be understood as a factor being highly random and indeterministic due to its ill-posed nature. In terms of SISR, we can define $\mu_{\text{true}}$ and $\epsilon$ with respect to an observed LR image $x$ and the corresponding HR image sample $y$ as following:
\begin{equation}
    \begin{aligned}
    \mu_\text{true} := \int yp(y|x) dy
    ,
    \end{aligned}
\end{equation}
\begin{equation}
        \begin{aligned}
        \epsilon := y - \mu_\text{true}
        ,
        \end{aligned}
\end{equation}
where, $\epsilon$ is expected to reside in high-frequency regions within every HR image sample, which makes exact pixel-wise reconstruction impossible.
Accordingly, representing the vanilla $L_1$ loss in terms of the components defined above, the original objective function can be reformulated as follows:
\begin{equation}
    \label{eq:original_loss}
    ||y^* - f(x)|| = ||\mu_{\text{true}} + \epsilon^* - f(x) ||,
\end{equation}
where $\epsilon^*$ is the inherent noise term for ground truth image $y^*$ in the training dataset. 
In the following sections, we will provide a comprehensive analysis based on this formulation.

% --------------------------------------------------------------------- %
% --------------------------------------------------------------------- %
% --------------------------------------------------------------------- %

\subsection{Modifying the objective function}
\paragraph{Taking the ill-posed nature into account.}
Regarding the ill-posed nature of SISR, multiple HR images can correspond to a single LR image. Therefore, following general principles in machine learning, it is natural to maximize the likelihood over all plausible solutions.
Accordingly, we begin our investigation by taking the posterior distribution into account and delve deeper into the underlying essentials of image super-resolution as below:
\begin{equation}
    \label{eq:posterior_basic}
    \begin{aligned}
          & \int ||y-f(x)||p(y|x)dy \\
        = & \int ||\mu_{\text{true}} + \epsilon - f(x) ||p(y|x)dy
        .
    \end{aligned}
\end{equation}
Then given an LR image $x$, an ideal SR model should estimate $\mu_{\text{true}}$, which is the optimal point of maximum likelihood, regarding that $\mathbb{E}_{p(y|x)}(y)=\mu_{\text{true}}$ by construction.

% \paragraph{Recap on the original L1 loss}
\paragraph{Vanilla training induces noisy training.}
It is worth noting that the original loss Eq.\eqref{eq:original_loss} is a specific case of Eq.\eqref{eq:posterior_basic}. If we let $p(y|x)$ as a Delta function where $p(y|x)=0$ for all points except for $y=y^*$, Eq.\eqref{eq:posterior_basic} is found to be identical to the original objective function.
Based on this observation, we can conclude that the current training protocol, indeed, fails in capturing the ill-posed nature of inverse problems. Instead, it treats the given HR sample as a unique and well-defined solution.  
However, this assumption does not account for the non-deterministic mapping from LR to HR, which makes the use of the Delta function for $p(y|x)$ less appropriate.
Moreover, this induces \textit{inherent noise} $\epsilon$ per every HR image, which can potentially hinder the stability of the training procedure.
However, in general, it is hard to disentangle the noise term since $\mu_{\text{true}}$ is intractable.
In further sections, we provide systematic methods to remove the noise term and enable optimization towards the centroid.

\section{Noise-free Objective Function}
\label{sec:noise_free}

\subsection{Removing the noise term}

In this section, our goal is to remove the inherent noise term in Eq.\eqref{eq:posterior_basic}, which can hinder the optimization, and only retain the centroid term. 
For any measurable and convex function $\phi(\cdot)$, we can obtain a lower bound of the expectation as $\mathbb{E}(\phi(\cdot)) \geq \phi(\mathbb{E(\cdot)})$ by Jensen's inequality. Since all $L_p\text{-norms}$ are convex for $p\geq1$; and $\mu_{\text{true}}$ and $f(x)$ are independent from $y$; and $\mathbb{E}_{y \sim p(y|x)}(\epsilon) = 0$ by definition, Eq.\eqref{eq:posterior_basic} can be simplified as following:
\begin{equation}
    \label{eq:lowerbound_true}
    \begin{aligned}
        & \mathbb{E}_{y \sim p(y|x)}(||\mu_{\text{true}} + \epsilon -f(x)||) 
        \\
        \geq & ||\mathbb{E}(\mu_{\text{true}}) + \mathbb{E}(\epsilon) - \mathbb{E}(f(x))|| 
        \\
        = & ||\mu_{\text{true}} - f(x)||.
    \end{aligned}
\end{equation}
By eliminating the \textit{per sample} inherent noise, we obtain a noise-free lower bound of the original objective function.
% , allowing healthy optimization towards the optimal maximum likelihood point.

%
% --------------------------------------------------------------------
%

\subsection{Empirical centroid estimation}
Although a noise-free objective has been obtained in Eq.\eqref{eq:lowerbound_true}, the true centroid term is still intractable and cannot be directly utilized since it involves taking the expectation over an infinite number of possible HR images.
Here, pretrained networks serve as a remedy to the problem at hand. It has been observed that low-level vision methods with pixel-wise loss implicitly tend to estimate the average among all plausible estimations \cite{coo_1, coo_2, SISR6_SRGAN, SISR7_ESRGAN} . This phenomenon, which we refer to as \textit{centroid-oriented optimization}, is acknowledged as a limitation of the training paradigm. 
However, by carefully integrating the retrospective centroid-oriented optimization phenomenon into the original training scheme in advance (i.e., by explicitly targeting the centroid), surprisingly, we can achieve favorable results. 
To this extent, we employ a pretrained super-resolution network as a centroid estimator. Thus, we refer to the estimation of a pretrained network as an \textit{empirical centroid}, which can be simply defined as follows:
\begin{equation}
    \mu_{\text{emp}} := \hat{f}(x),
\end{equation}
where $\hat{f}$ is the pretrained SR network. 
Here, the empirical centroid $\mu_{\text{emp}}$ can be understood as the expectation, but with regard to the learned natural image prior obtained by the training dataset of the pretrained network.

\section{Estimation Error of Empirical Centroids}

In the previous section, we leveraged a pretrained network as an approximation of the centroid of the posterior distribution. However, even the state-of-the-art pretrained networks are followed by estimation errors, and thus should not be treated as ideal networks. 
Here, we examine the estimation errors from the perspectives of both (1) low-frequency (LF) components, which can be observed when SR images do not downsample to the original LR images, and (2) high-frequency (HF) components, which are the case when SR images only contain limited sharp details, below the theoretical upper-bound of pixel-wise reconstruction. Hence, we start this section by reformulating Eq.\eqref{eq:lowerbound_true} as following:
\begin{equation}
    \label{eq:lowerbound_with_estimationerror}
    ||(\mu_{\text{emp}} + \Delta\mu) - f(\downarrow(\mu_{\text{emp}} + \Delta\mu + \epsilon))||,
\end{equation}
where $\Delta\mu:=\mu_{\text{true}}-\mu_{\text{emp}}$ is the estimation error and $\downarrow$ is the downsampling operation.
We emphasize that these limitations of pretrained networks should be taken into account, which will be further discussed in the following sections.

\paragraph{Revisiting Knowledge Distillation.}
Here, we demonstrate that a well-known training technique, Knowledge Distillation (KD), can be simply represented in terms of the components derived in the previous sections as below:
\begin{equation}
        \label{eq:construction_of_kd}
        \begin{aligned}
        &||\hat{f}(x) - f(x)|| \\
        =&||\mu_{\text{emp}} - f(x)|| \\
        =&||(\mu_{\text{emp}} + \Delta\cancel{\mu}) - f(\downarrow(\mu_{\text{emp}} + \Delta\mu + \epsilon))||
        ,
        \end{aligned}
\end{equation}
where the first row is the original formulation of KD and the others are equivalent objectives in terms of our observation. 
This can be understood as a special case of Eq.\eqref{eq:lowerbound_with_estimationerror}, with $\Delta \mu = 0$ only on the left term. In other words, the objective of KD (Eq.\eqref{eq:construction_of_kd}) neglects the estimation error of the teacher model in the target image but leaves it in the LR image. 
However, predictions of pretrained networks may not downsample to the original LR image precisely due to the LF components of 
$\Delta \mu$, and conversely, the given HR image will not align with the corresponding LR image.
We refer to this discrepancy as spatial inconsistency between the input and target images, highlighting a critical limitation in the formulation of KD.
Specifically, this spatial inconsistency hinders KD to provide proper supervision, thereby leading to potential instability in the training process.
Additionally, since the estimation error term $\Delta \mu$ of the target image is ignored, this term will not be optimized which leads to limited performance bounded by the teacher network.
Overall, while KD-based training may benefit from the noise-free objective and converge faster in the early steps of training, it will suffer from additional challenges by ignoring $\Delta \mu$ only in the target image.

\section{Empirical Centroid-oriented Optimization} \label{sec:ecoo_method}

%Previous approach on modeling a noise-free optimization objective was followed by apparent flaws.
In this section, we make a quick fix on the limitations of conventional KD observed above. We construct a noise-free optimization objective in a spatially consistent manner, followed by a method to handle the estimation error. 

\subsection{Spatially consistent noise-free objective}
Regarding that $\mu_{\text{true}}$ are linear combinations of plausible HR images, $f(\downarrow(y^*))=f(\downarrow(\mu_{\text{true}}))$ holds if the network $f$ and the downsampling operation $\downarrow$ are linear. By taking into account the piece-wise linearity \cite{PiecewiseLinear} of $f$ and the fact that ``plausible'' HR images downsample to identical images by construction, we make a fair approximation of Eq.\eqref{eq:lowerbound_true} as follows:
\begin{equation}
    \label{eq:lowerbound_withlinearity_}
    \begin{aligned}
        ||\mu_{\text{emp}} + \Delta\mu - f(\downarrow(\mu_{\text{emp}} + \Delta\mu))||
        .
    \end{aligned}
\end{equation}
Instead of assuming $\Delta\mu=0$ only on the left side as in KD, we remove $\Delta\mu$ in \textbf{both} terms of the approximation and propose an objective as below:
\begin{equation}
    \label{eq:train_with_sr}
    \begin{aligned}
        &||\mu_{\text{emp}} + \cancel{\Delta\mu} - f(\downarrow(\mu_{\text{emp}} + \cancel{\Delta\mu}))|| \\
         =&||(\mu_{\text{emp}}) - f(\downarrow(\mu_{\text{emp}})||
         .
    \end{aligned}
\end{equation} 
This way, we obtain a tractable noise-free objective function, which enables the proposed Empirical Centroid-oriented Optimization (ECO) without risking the optimization procedure from spatial inconsistency observed in KD.  

\subsection{Taking the estimation error into account}\label{section:mixup_as_rescue}
\paragraph{Trade-off of removing the error term.}
\label{section:trade-off of approximated noise-free objective}
While it is important to prevent highly random and noisy HF components from disturbing the training, removing more HF components than required (i.e., over-smoothing) will lead to failure in providing sufficient supervision for necessary detail recovery. 
Regarding that pretrained networks can fail to generate sharp details, the problem of insufficient HF supervision still remains in the objective in Eq.\eqref{eq:train_with_sr}.
Thus, Eq.\eqref{eq:train_with_sr} has a trade-off between stable training and the limited capability of HF supervision. In practice, the impact of neglecting $\Delta\mu$ can empirically be larger than the benefit of noise-free objective after sufficient training iterations, where networks need to be fine-tuned. Overall, both our tractable noise-free objective Eq.\eqref{eq:train_with_sr} and the vanilla training objective Eq.\eqref{eq:original_loss} come with their own set of advantages and disadvantages. 

% The noise-free objective provides faster optimization and stability during early training, but it suffers from over-smooth target images, 
% between $\mu_{\text{true}}$ and $\mu_{\text{emp}}$, especially when longer training is involved. Meanwhile, vanilla training suffers from inherent noise residing in very high-frequency regions in early training steps. 

\paragraph{Mixup as rescue.}
To this extent, we propose a simple and efficient workaround to capture the advantages of both Eq.\eqref{eq:train_with_sr} and Eq.\eqref{eq:original_loss}. The proposed method starts by training the network with our tractable noise-free objective in Eq.\eqref{eq:train_with_sr}. However, once adequate convergence is achieved, we switch the objective to the original objective Eq.\eqref{eq:original_loss} and obtain additional supervision on HF components. 
Remarkably, it turns out that this type of approach can be formulated with a well-known data augmentation method, \textit{mixup} \cite{mixup}. As the first step, we reformulate the original loss function Eq.\eqref{eq:original_loss} as follows:
\begin{equation}\label{eq:original_loss_in_mixupform}
    \begin{aligned}
        &||y^* - f(\downarrow(y^*))|| \\
        =&||(\mu_{\text{true}} + \epsilon) - f(\downarrow(\mu_{\text{true}} + \epsilon))|| \\
        =&||(\mu_{\text{emp}}+ 1(\Delta\mu+\epsilon)) - f(\downarrow(\mu_{\text{emp}} + 1(\Delta\mu+\epsilon))||. \\
    \end{aligned}
\end{equation}
Equally, the objective function based on mixup can be interpreted as an additive term of a single data pair and another as below:
\begin{equation}
    \begin{aligned}
        &L(\alpha Y_1 + (1-\alpha) Y_2, \phi(\alpha X_1 + (1-\alpha) X_2)) \\
        =&L(Y_2 + \alpha (Y_1 - Y_2), \phi(X_2 + \alpha (X_1 - X_2)), \\
    \end{aligned}
\end{equation}
where $L(\cdot,\cdot)$ is an arbitrary loss function with inputs $X_1, X_2$, targets $Y_1, Y_2$ and the network to optimize as $\phi$. Here, if we let $L(\cdot,\cdot)$ as the pixel-wise norm, $\phi=f$, $(X_1, Y_1)$ as the original data pair $(x, y^*)$ and $(X_2, Y_2)$ as the synthetic data pair $(\downarrow(\mu_\text{emp}),\mu_\text{emp})$, we can obtain our final objective function as follows:
\begin{equation}
    \label{eq:final_w_mixup}
    \begin{aligned}
        &||(\mu_{\text{emp}}+ \alpha(y^* - \mu_{\text{emp}})) - f(\downarrow(\mu_{\text{emp}}+ \alpha(y^* - \mu_{\text{emp}})))|| \\
        =&||(\mu_{\text{emp}}+ \alpha(\Delta\mu+\epsilon^*)) - f(\downarrow(\mu_{\text{emp}} + \alpha(\Delta\mu+\epsilon^*)))||
        .
    \end{aligned}
\end{equation}
With a smooth transition of $\alpha=0$ to $\alpha=1$, we can easily balance through the spatially aligned tractable noise-free objective ($\alpha=0$) and the vanilla objective ($\alpha=1$).
% two objectives aforementioned (Eq.\eqref{eq:train_with_sr}, Eq.\eqref{eq:original_loss_in_mixupform}). 
%
It should be noted that the inherent noise will be reintroduced back into the training as $\alpha$ increases.
%
% However, our empirical findings (Sec.\ref{section:experiments}) are that the networks are sufficiently stabilized in later steps, capable of tolerating the reintroduced noise while they benefit from enhanced HF supervision.
%
However, our empirical findings in Sec.\ref{section:experiments} reveal that the early stages of training play a crucial role in overall performance. In later steps, networks become relatively stabilized, allowing them to tolerate the reintroduced noise while benefiting from enhanced high-frequency (HF) supervision.
Overall, this balanced approach allows for the advantages of noise-free training in the early stages without sacrificing the benefits of HF supervision in later training.
By preprocessing synthetic images and parallelizing mixup with separate CPU processes, the proposed method can be implemented in just a few lines of code.
The overall framework of our method is illustrated in Fig.\ref{fig:main_figure}. Unless specified otherwise, the term `ECO' throughout this paper refers to our proposed method together with the usage of the mixup strategy described in Eq.\eqref{eq:final_w_mixup}.

\paragraph{Difference with conventional mixup.} 
% It is important to 
% Here, we clarify the difference between our method against conventional mixup. 
The proposed method is a mixture of the original (HR, LR) image pairs and synthetic reconstruction of the \textit{identical} images. On the other hand, conventional mixup refers to blending between \textit{different} data samples in order to augment limited data samples. 
% Another important difference is that, unlike conventional mixup, we have a scheduling hyperparameter namely $\alpha$, which will be gradually increased from 0 to 1, as training continues. 
Note that these two methods are fairly orthogonal and can be applied simultaneously.

\section{Experiments}\label{section:experiments}
\subsection{Analyzing the impact of noise-free training} \label{section:analysis}
We use EDSR-baseline \cite{SISR2_EDSR} as the representative model and investigate the impact of the noise-free objective obtained in Eq.\eqref{eq:train_with_sr}, without mixup.

\begin{figure}[t]
    \begin{center}
    \includegraphics[width=\columnwidth]{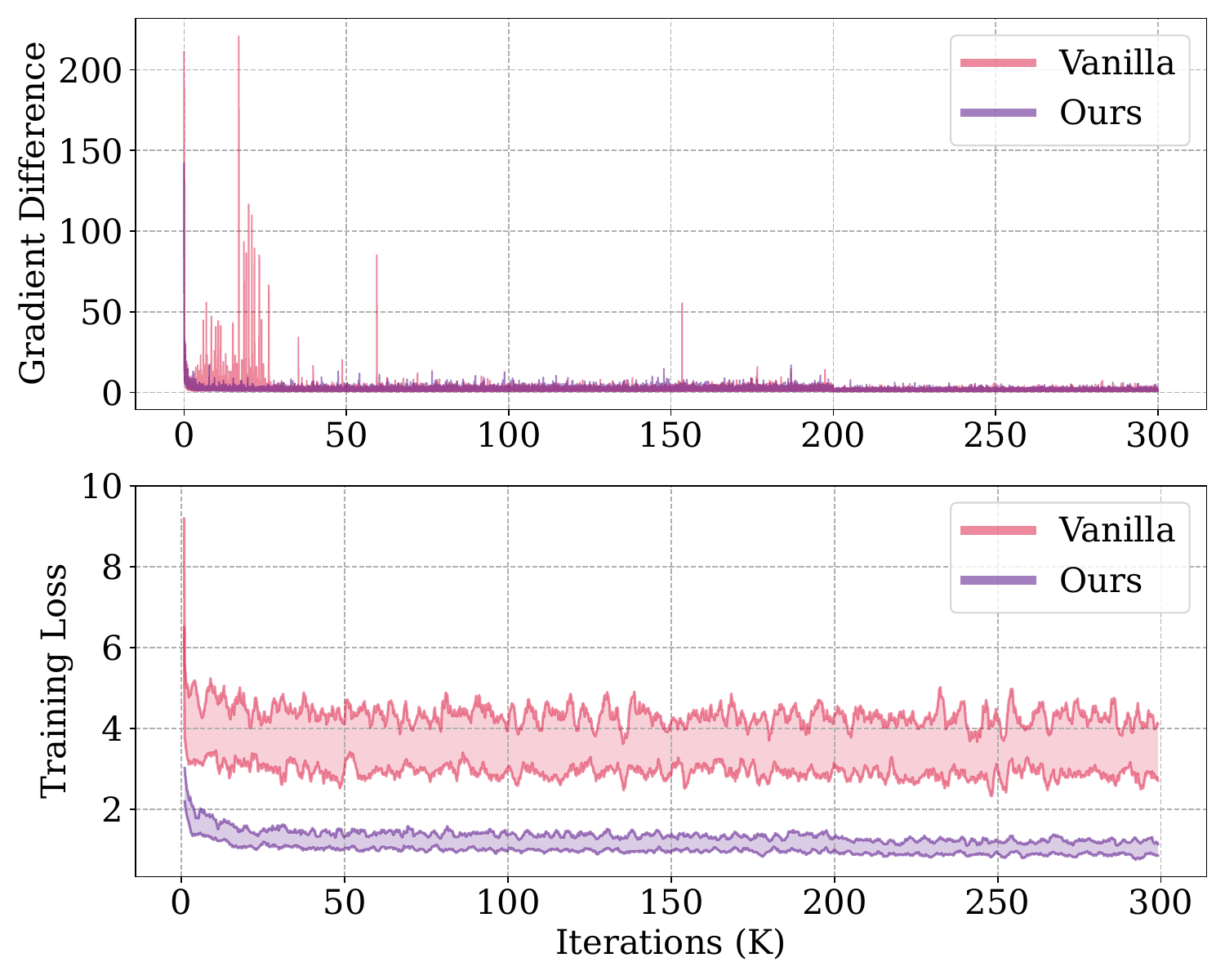}
    \end{center}
    \vspace{-10pt}
    \caption{
    Visualization of maximum gradient difference and the loss variation. Spikes of gradient differences indicate that the gradients are not well-bounded (i.e., not Lipschitz).
    }
    \label{fig:optimization_landscape}
\end{figure}

\paragraph{Exploring the optimization landscape.}
Following \cite{how_does_batchnorm_help}, we identify the impact of the proposed noise-free objective within the training process by investigating the optimization landscape and the Lipschitzness of the loss function.
At each specific training point, we move through the gradient direction and observe the loss variation and the maximum gradient difference in terms of $L_2$-norm, as illustrated in Fig.\ref{fig:optimization_landscape}.
Through the use of the noise-free objective, we observe well-bounded loss values, which aligns with our theoretical analysis. 
Moreover, of greater importance is that while vanilla training leads to sharp spikes during early training steps, noise-free training shows well-bounded gradients.
In other words, noise-free training demonstrates a notably improved level of effective $\beta$-smoothness \cite{beta_smoothness, how_does_batchnorm_help}. 
In the context of gradient-based training methods, it is clear that the overall training procedure can be significantly influenced by gradient behaviors. Specifically, vanishing or exploding gradients can raise additional challenges when training deep networks.
Thus, by having a well-behaving and predictable gradient with the proposed noise-free objective, we can alleviate these issues and obtain faster convergence with improved stability.
%
% -------------- Important!
This observation underlines the significance of noise-free training during the early stages, as it minimizes fluctuations and instabilities that could hinder the learning process. By enhancing stability in these crucial initial steps, our method can lead to an overall performance gain, setting a strong foundation for later stages of training.
%
%
%
% 
% Further, since the noise-free optimization landscape is shown to be smoother, we can minimize the risk of falling into sudden local minima in the early stages which can lead to an overall performance drop.
%

\begin{figure}[t]
    \begin{center}
    \includegraphics[width=\columnwidth]{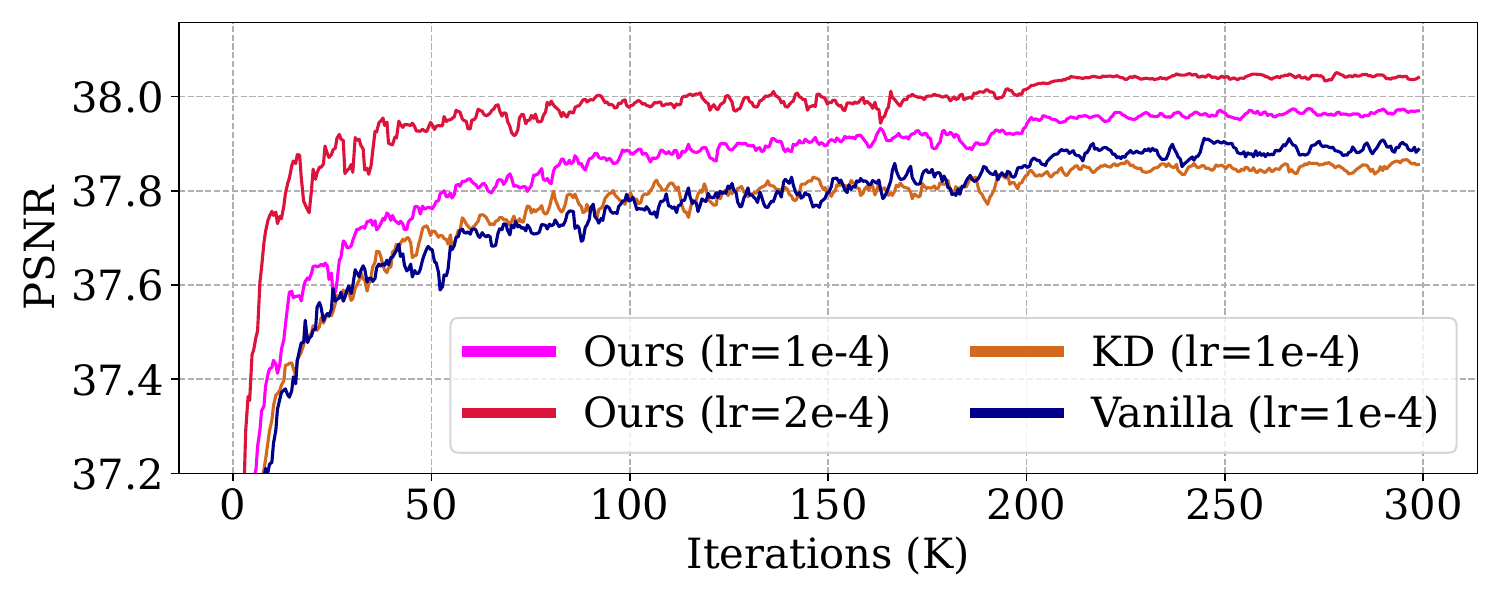}
    \end{center}
    \vspace{-10pt}
    \caption{
    Comparison of our method (w/o mixup) with KD and vanilla training on Set5. It verifies the impact of spatial inconsistency in training image pairs.
    }
    \vspace{-5pt}
    \label{fig:different_training_schemes}
\end{figure}
\paragraph{Comparison against vanilla training and KD.}
In Fig.\ref{fig:different_training_schemes}, we provide training curves of noise-free training (w/o mixup) against vanilla training and knowledge distillation (KD). 
% Following \cite{SISR2_EDSR}, we ado pt the EDSR-baseline network with 300K training iterations for all training schemes. 
%%
It demonstrates that KD can also lead to slightly faster convergence during early training since the formulation of KD is also expected to have noise-free targets. However, we have shown that it is followed by a fundamental limitation: spatial inconsistency between input and target images. Accordingly, the final performance turns out to be worse than that of vanilla training, while the proposed spatially aligned noise-free objective obtains overall performance gain. 
Remarkably, despite the only changes being the construction of LR images, it shows significant improvement.

% -- KD training experiment

\begin{figure}[h]
    \includegraphics[width=\columnwidth]{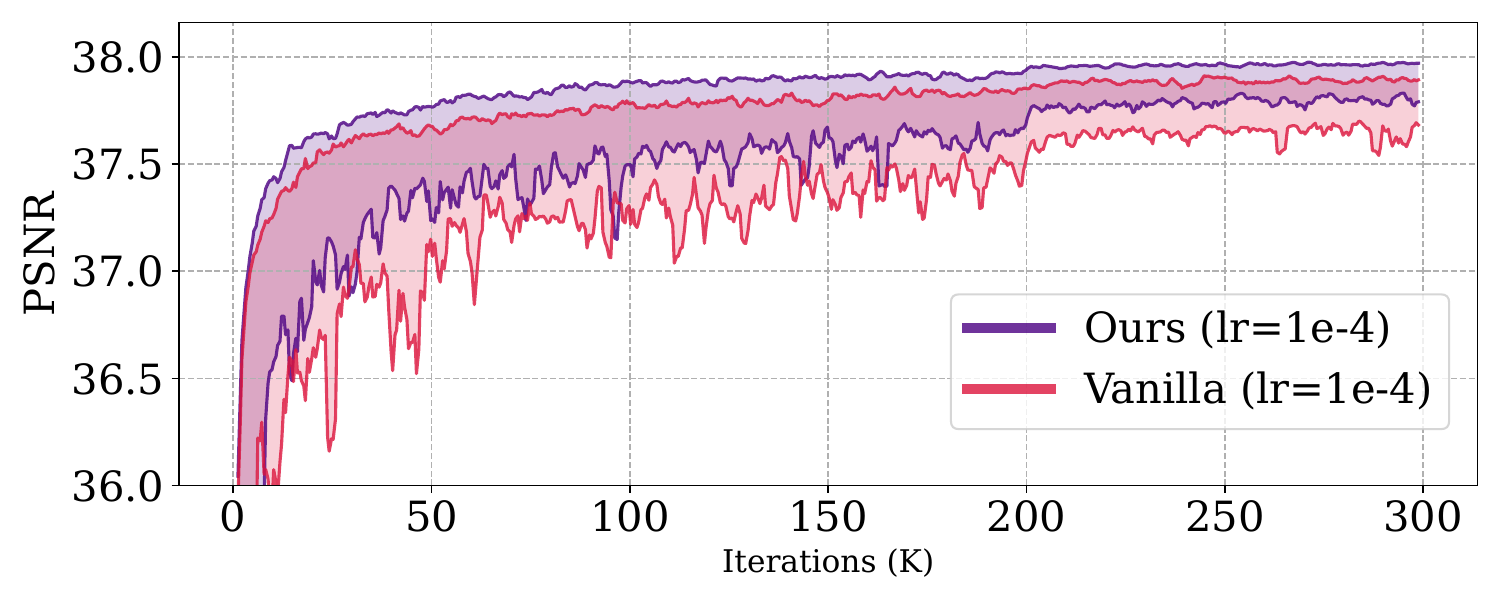}
    \caption{
    Validation results are reported for both vanilla training and the proposed method (without mixup) across mini-batch sizes of 2, 4, 8, and 16. The shaded regions indicate the minimum and maximum PSNR values at each iteration across all settings. Noise-free optimization enables additional stability throughout various batch-size choices.
    }
    \label{fig:different_batch_size}
\end{figure}

\paragraph{Comparison over various batch-size.}
With smaller mini-batch sizes, each gradient step becomes more reliant on every individual data point within the batch. In the case of vanilla training, the training procedure becomes more susceptible to \textit{per sample} noise originating from each image instance. 
Comparatively, the proposed method is relatively free from per-sample noise, which enables additional robustness to smaller mini-batch size selection. To validate the statement, we perform extensive experiments over various selections of smaller mini-batch sizes as in Fig.\ref{fig:different_batch_size}. 
% We adopt the widely used EDSR-baseline \cite{SISR2_EDSR} network and compare vanilla training against our method. 
The mini-batch size is chosen as 2, 4, 8, and 16 where 16 is the default setting for most works. 
As demonstrated in Fig.\ref{fig:different_batch_size}, vanilla training shows fluctuating PSNR scores with small mini-batch sizes, especially in early training steps, while our method provides increased stability and faster convergence over various mini-batch size choices. 

\begin{figure}[h]
    \includegraphics[width=\columnwidth]{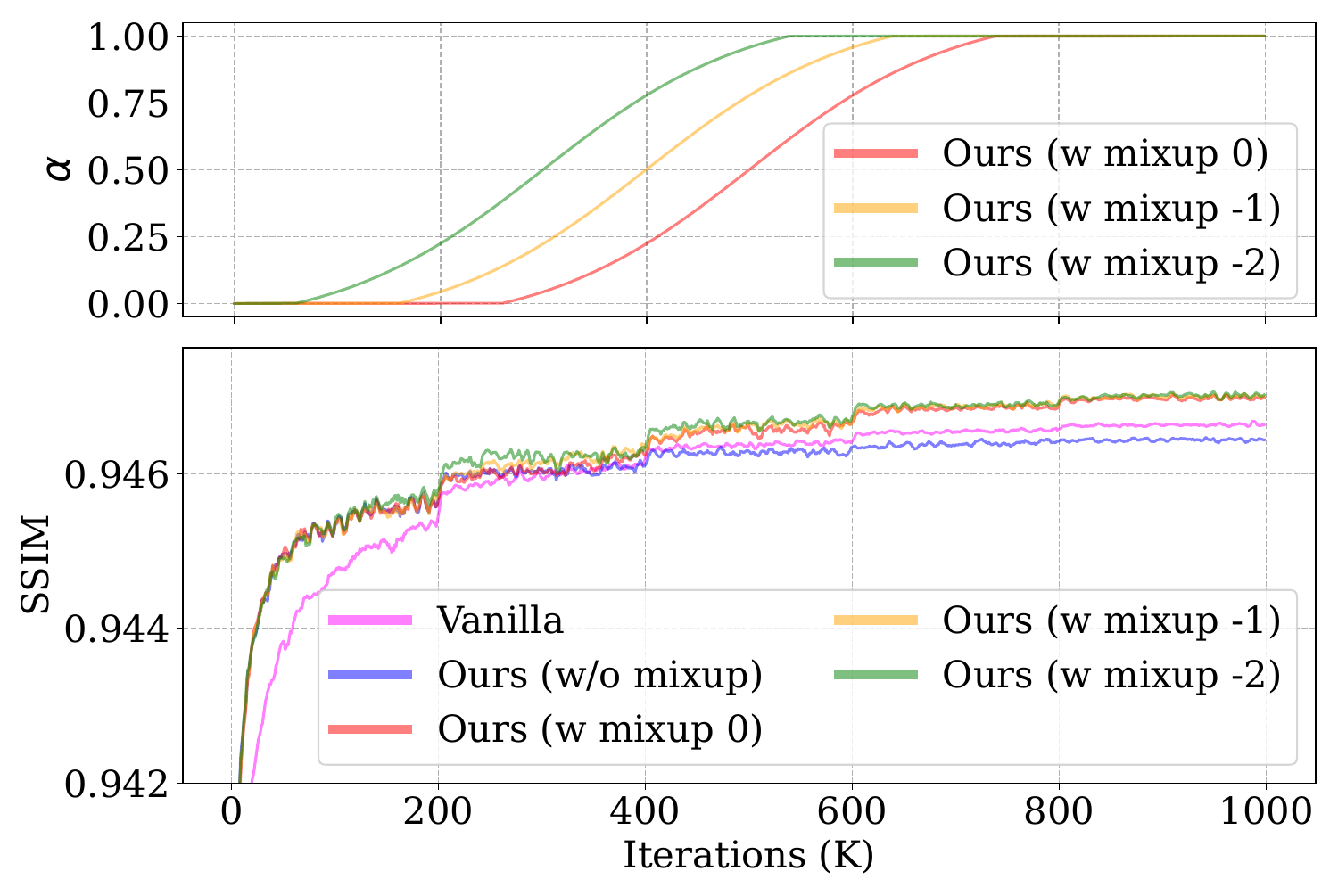}
    \caption{
    Validation results over various configurations of mixup. Without mixup, the performance is limited due to neglecting the estimation error factor $\Delta\mu$ as in Eq.\eqref{eq:train_with_sr}. 
    }
    \label{fig:longer_training}
\end{figure}

\paragraph{Empirical impact of the estimation error.}
% -- Longer training & Empirical Impact
Fig.\ref{fig:longer_training} illustrates the empirical trade-off between smoother gradients and the ignorance of the estimation error. In the early stages, we can observe clear improvement when training with the proposed noise-free objective. However, the impact of the estimation error empirically increases, and the final performance turns out to be lower than that of the original training scheme if the mixup strategy is not used. Together with mixup, it is shown that we can obtain superior performance over the entire training steps. We have further analyzed the mixup strategy by shifting the scheduling hyperparameter $\alpha$ in Eq.\eqref{eq:final_w_mixup} but did find it to be significant.
%
%%%%%%%%%%%%%%%%%%%%%%%%%%%%%%%%%%%%%%%%%%%%%%%%%%%%%%%%%%%%%%%%%%%%%%%%%%%%%%%%%%%%%%%%%%%%%%%
%%%%%%%%%%%%%%%%%%%%%%%%%%%%%%%%%%%%%%%%%%%%%%%%%%%%%%%%%%%%%%%%%%%%%%%%%%%%%%%%%%%%%%%%%%%%%%%
%
%
%

\begin{table*}[!t]
\begin{center}
\resizebox{\textwidth}{!}{
\small
\begin{tabular}{c|l|c|c|c|c|c|c}
\hline
Scale & \multicolumn{1}{c|}{Model} & \multicolumn{1}{c|}{Method} & Set5 & Set14 & BSD100 & Urban100 & Manga109 \\
\hline\hline
\multirow{4}{*}{$\times$2}
%& EDSR & \multicolumn{1}{l}{Vanilla}-16     & 99.99 / 0.9999 & 99.99 / 0.9999 & 99.99 / 0.9999 & 99.99 / 0.9999 & 99.99 / 0.9999 \\
& EDSR \cite{SISR2_EDSR} & \multicolumn{1}{l|}{Vanilla}    & 38.18 / 0.9612 & 33.82 / 0.9197 & 32.33 / 0.9016 & 32.83 / 0.9349 & 39.05 / 0.9777 \\
& EDSR \cite{SISR2_EDSR} & \multicolumn{1}{l|}{ECO (ours)} & \textbf{38.29} / \textbf{0.9615} & \textbf{34.07} / \textbf{0.9210} & \textbf{32.37} / \textbf{0.9022} & \textbf{33.07} / \textbf{0.9369} & \textbf{39.26} / \textbf{0.9782} \\
\cline{2-8}
%& RCAN & \multicolumn{1}{l}{Vanilla}-16     & 99.99 / 0.9999 & 99.99 / 0.9999 & 99.99 / 0.9999 & 99.99 / 0.9999 & 99.99 / 0.9999 \\
& RCAN \cite{SISR4_RCAN} & \multicolumn{1}{l|}{Vanilla}    & 38.26 / \textbf{0.9615} & 34.04 / \textbf{0.9215} & 32.35 / 0.9019 & 33.05 / 0.9364 & 39.34 / \textbf{0.9783} \\
& RCAN \cite{SISR4_RCAN} & \multicolumn{1}{l|}{ECO (ours)} & \textbf{38.28} / \textbf{0.9615} & \textbf{34.07} / \textbf{0.9215} & \textbf{32.39} / \textbf{0.9023} & \textbf{33.22} / \textbf{0.9378} & \textbf{39.39} / \textbf{0.9783} \\
\hline\hline
\multirow{4}{*}{$\times$3}
%& EDSR & \multicolumn{1}{l}{Vanilla}-16     & 99.99 / 0.9999 & 99.99 / 0.9999 & 99.99 / 0.9999 & 99.99 / 0.9999 & 99.99 / 0.9999 \\
& EDSR \cite{SISR2_EDSR} & \multicolumn{1}{l|}{Vanilla}    & 34.70 / 0.9294 & 30.58 / 0.8468 & 29.26 / 0.8095 & 28.75 / 0.8648 & 34.17 / 0.9485 \\
& EDSR \cite{SISR2_EDSR} & \multicolumn{1}{l|}{ECO (ours)} & \textbf{34.80} / \textbf{0.9302} & \textbf{30.64} / \textbf{0.8476} & \textbf{29.32} / \textbf{0.8108} & \textbf{28.95} / \textbf{0.8679} & \textbf{34.36} / \textbf{0.9496} \\
\cline{2-8}
%& RCAN & \multicolumn{1}{l}{Vanilla}-16     & 99.99 / 0.9999 & 99.99 / 0.9999 & 99.99 / 0.9999 & 99.99 / 0.9999 & 99.99 / 0.9999 \\
& RCAN \cite{SISR4_RCAN} & \multicolumn{1}{l|}{Vanilla}    & 34.80 / 0.9302 & 30.62 / 0.8476 & 29.32 / 0.8107 & 29.01 / 0.8685 & 34.48 / 0.9500 \\
& RCAN \cite{SISR4_RCAN} & \multicolumn{1}{l|}{ECO (ours)} & \textbf{34.86} / \textbf{0.9306} & \textbf{30.68} / \textbf{0.8484} & \textbf{29.33} / \textbf{0.8111} & \textbf{29.09} / \textbf{0.8700} & \textbf{34.56} / \textbf{0.9504} \\
\hline
\hline
\multirow{4}{*}{$\times$4}
%& EDSR & \multicolumn{1}{l}{Vanilla}-16     & 99.99 / 0.9999 & 99.99 / 0.9999 & 99.99 / 0.9999 & 99.99 / 0.9999 & 99.99 / 0.9999 \\
& EDSR \cite{SISR2_EDSR} & \multicolumn{1}{l|}{Vanilla}    & 32.50 / 0.8986 & 28.81 / 0.7871 & 27.71 / 0.7416 & 26.55 / 0.8018 & 30.97 / 0.9145 \\
& EDSR \cite{SISR2_EDSR} & \multicolumn{1}{l|}{ECO (ours)} & \textbf{32.59} / \textbf{0.8998} & \textbf{28.90} / \textbf{0.7892} & \textbf{27.78} / \textbf{0.7432} & \textbf{26.77} / \textbf{0.8064} & \textbf{31.32} / \textbf{0.9182} \\
\cline{2-8}
%& RCAN & \multicolumn{1}{l}{Vanilla}-16     & 99.99 / 0.9999 & 99.99 / 0.9999 & 99.99 / 0.9999 & 99.99 / 0.9999 & 99.99 / 0.9999 \\
& RCAN \cite{SISR4_RCAN} & \multicolumn{1}{l|}{Vanilla}    & \textbf{32.71} / 0.9008 & 28.87 / 0.7887 & 27.77 / 0.7434 & 26.83 / 0.8078 & 31.31 / 0.9168 \\
& RCAN \cite{SISR4_RCAN} & \multicolumn{1}{l|}{ECO (ours)} & 32.70 / \textbf{0.9011} & \textbf{28.91} / \textbf{0.7895} & \textbf{27.80} / \textbf{0.7437} & \textbf{26.88} / \textbf{0.8086} & \textbf{31.38} / \textbf{0.9174} \\
\hline
\hline
\end{tabular}
}
\caption{Quantitative comparison of the proposed method ECO (w/ mixup) against vanilla training. We report PSNR (dB) and SSIM scores for $\times$2, $\times$3, and $\times$4 SR over standard benchmark datasets. The best result are highlighted in \textbf{bold}.}
\label{tab:maintable}
\end{center}
\end{table*}

\begin{figure*}[t]
    \includegraphics[width=\textwidth]{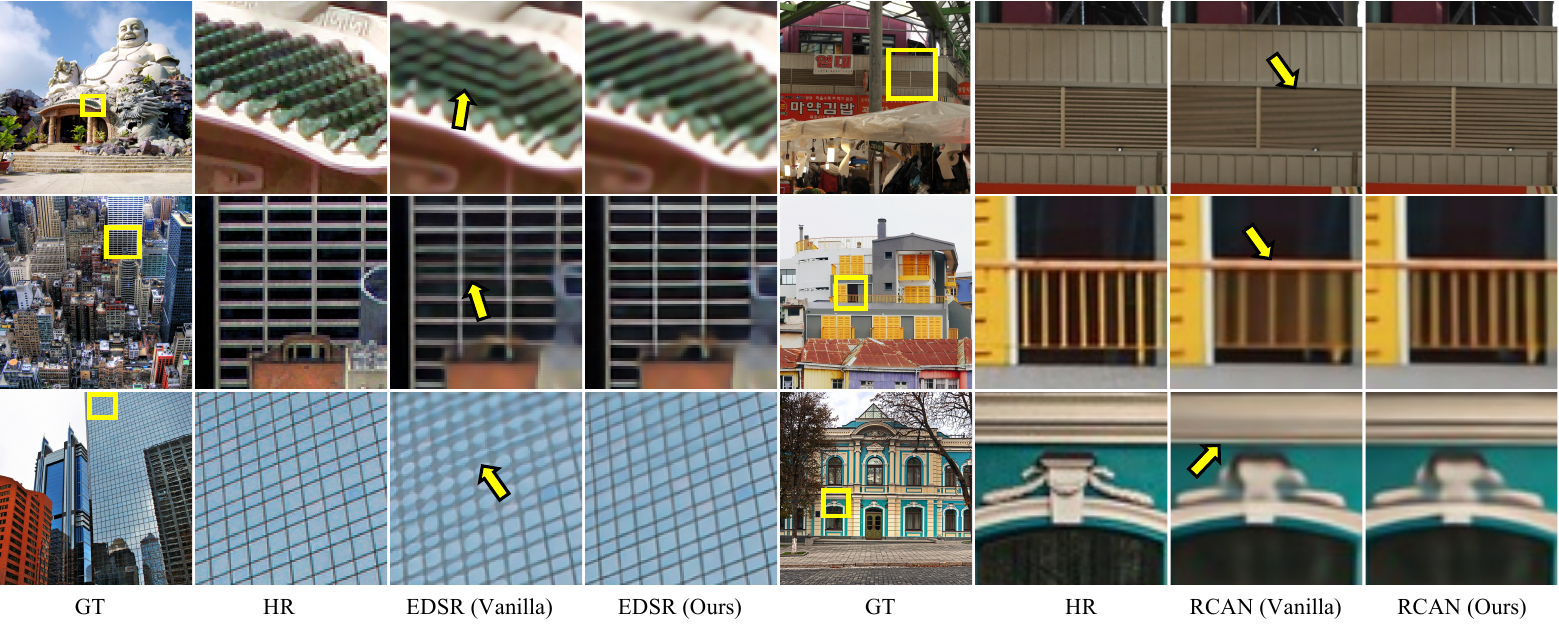}
    \caption{
    Visual comparison of the proposed method and vanilla training for $\times$4 SR. \textbf{Zoom in for best view.}
    }
    \label{fig:qualitative_result}
\end{figure*}

% \usepackage{booktabs}
\begin{table}[h]
\setlength\tabcolsep{4.0pt}
\footnotesize
\begin{tabular}{l|l|l|l|l}
\hline
\multicolumn{1}{c}{} & \multicolumn{1}{|c}{Set5} & \multicolumn{1}{|c}{Set14} & \multicolumn{1}{|c}{Urban100} & \multicolumn{1}{|c}{Manga109} \\ 
\hline
\hline
\multicolumn{5}{l}{\textbf{(a) $\times$2 SR on SwinIR-small}}\\
\hline
Vanilla             & 38.19/\textbf{.9613}             & 33.93/.9203             & 32.74/.9338             & 39.11/\textbf{.9781}                 \\
ECO               & \textbf{38.21}/\textbf{.9613}             & \textbf{33.96}/\textbf{.9209}             & \textbf{32.78}/\textbf{.9345}             & \textbf{39.16}/\textbf{.9781}                     \\ 
\hline
\hline
\multicolumn{5}{l}{\textbf{(b) $\times$2 SR on EDSR-baseline with L2 loss}}\\
\hline
Vanilla             & 37.89/.9601             & \textbf{33.47}/.9167             & 31.74/.9246             & 38.12/.9759                 \\
ECO               & \textbf{37.94}/\textbf{.9602}             & \textbf{33.47}/\textbf{.9170}             & \textbf{31.77}/\textbf{.9249}             & \textbf{38.34}/\textbf{.9764}                 \\ 
\hline
\hline
\multicolumn{5}{l}{\textbf{(c) $\times$2 SR on EDSR-baseline, on real-world dataset}}                                                                                                                                   \\
\hline
Vanilla             & 33.46/.9074             & 30.58/.8412             & 29.37/.8744             & 34.08/.9399                  \\
ECO              & \textbf{33.49}/\textbf{.9078}             & \textbf{30.60}/\textbf{.8416}             & \textbf{29.38}/\textbf{.8748}             & \textbf{34.16}/\textbf{.9403}                 \\ 
\hline
\hline
\multicolumn{5}{l}{\textbf{(d) $\times$8 SR on EDSR-baseline}}\\
\hline
Vanilla               & 26.88/\underline{.7712}             & 24.85/.6370             & 22.30/.6089             & 24.34/.7696                  \\  
ECO*              & \underline{26.90}/.7700                          & \underline{24.91}/\underline{.6378}             & \underline{22.37}/\underline{.6091}             & \underline{24.40}/\underline{.7697}                 \\ 
ECO              & \textbf{27.00}/\textbf{.7743}             & \textbf{24.94}/\textbf{.6398}             & \textbf{22.41}/\textbf{.6132}             & \textbf{24.52}/\textbf{.7749}                 \\
\hline
\hline
\end{tabular}
\caption{ECO (ours) compared to vanilla training. PSNR (dB) and SSIM scores are reported, and the best and second-best results are highlighted in \textbf{bold} and \underline{underlines}. ECO* indicates training only up to 20$\%$ of the total iterations.}
\label{tab:ablation_table}
\end{table}

\subsection{Evaluation on the state-of-the-art methods}

\paragraph{Experimental Setup.}
We validate the effectiveness of our method on benchmark datasets: Set5 \cite{set5}, Set14 \cite{set14}, BSD100 \cite{bsd100}, Urban100 \cite{urban100} and Manga109 \cite{manga109}. 
We reproduce all methods and mixup is used for our method. For Tab.\ref{tab:maintable}, we follow \cite{rcanit} and train networks with larger mini-batch size and fewer iterations in order to reduce the overall training time. See the supplementary materials for details.

\paragraph{Benchmark comparison.}
In Tab.\ref{tab:maintable}, we compare the proposed training scheme against vanilla training in standard SR settings. Specifically, evaluation is performed for $\times$2, $\times$3 and $\times$4 SR tasks with bicubic downsampling. It demonstrates that our method leads to sustainable performance gain in terms PSNR and SSIM over standard benchmark datasets. In qualitative comparison (Fig.\ref{fig:qualitative_result}) for $\times 4$ SR, we can clearly see that the proposed method provides more visually pleasing results, successfully recovering high-frequency details. 

\paragraph{Larger scale factor and adaptation to real-world.}
We further perform extensive experiments comparing our method against vanilla training in $\times$8 SR task and real-world $\times$2 SR settings. 
In the case of the real-world setting, LR images with additive color Gaussian noise were used for both training and evaluation and the average score of 10 different runs is reported. 
Tab.\ref{tab:ablation_table}.(c) and Tab.\ref{tab:ablation_table}.(d)
indicate that the proposed training framework leads to performance gain in both real-world $\times$2 SR and bicubic $\times$8 SR. Remarkably, we reach comparable performance to vanilla training with only 20$\%$ of the total iterations for $\times$8 SR. It verifies the higher benefits of noise-free training when the inherent noise term is expected to exhibit greater randomness.

\paragraph{Independence of architecture and loss.}
In Tab.\ref{tab:ablation_table}.(a-b), we further validate the proposed training framework with SwinIR \cite{liang2021swinir} and with the $L_2$ loss, respectively.
Experimental results verify that the application of the proposed method is not limited to only CNN architectures or the L1 loss.
% which aligns with our theoretical? construction since we have not assumed any specific network architecture or $L_p$-norm to obtain .

%%%%%%%%%%%%%%%%%%%%%%%%%%%%%%%%%%%%%%%%%%%%%%%%%%%%%%%%%%%%%%%%%%%%%%%%%%%%%%%%%%%%%%%%%%%%%%%%%
%%%%%%%%%%%%%%%%%%%%%%%%%%%%%%%%%%%%%%%%%%%%%%%%%%%%%%%%%%%%%%%%%%%%%%%%%%%%%%%%%%%%%%%%%%%%%%%%%
%%%%%%%%%%%%%%%%%%%%%%%%%%%%%%%%%%%%%%%%%%%%%%%%%%%%%%%%%%%%%%%%%%%%%%%%%%%%%%%%%%%%%%%%%%%%%%%%%

% --- mixup effect
\section{Related Work}
%

% CNN based
Starting with the pioneering work \cite{SISR1_SRCNN}, CNN base networks \cite{SISR5_SAN, SISR12_HAN, SISR2_EDSR, SISR3_VDSR, SISR4_RCAN, SISR11_CRAN} aiming for high fidelity reconstruction has shown drastic development.
%
% ViT Swin based
Later, ViT and Swin-based networks \cite{chen2021pre_transformer1, liang2021swinir, zhang2022swinfir, chen2023hat} have achieved the state-of-the-art performances revealing the effectiveness of self-attention in context of image reconstruction.
Several works investigate the objective function of SISR \cite{he2022revisiting_illposed_2, ning2021uncertainty} and empirical results of \cite{SISR2_EDSR} demonstrate that the $L_1$ loss can lead to better convergence against the widely used $L_2$ loss.
Knowledge distillation methods \cite{zhang2021data_distillation1, wang2021towards_distillation2, lee2020learning_distillation3, gao2019image_distillation4} have shown their efficiency on small SR networks where \cite{lee2020learning_distillation3} uses privileged information to boost the teacher network's performance.
Meanwhile, \cite{lew2021pixel, ikc, kernelgan} aims to model the complex degradations explicitly. 
%
% Meanwhile, empirical results of \cite{SISR2_EDSR} demonstrate that the $L_1$ loss can lead to better convergence against the widely used $L_2$ loss, and \cite{fuoli2021fourier} introduces Fourier domain losses as an efficient objective function.
%
%
To tackle the ill-posed nature of SISR, several methods \cite{SISR6_SRGAN, SISR7_ESRGAN, SISR8_RankSRGAN} obtain enhanced visual quality by utilizing the adversarial loss and the perceptual loss \cite{perceptual_loss_vgg}. Further, \cite{jo2021tackling_illposed_1} generates adaptive targets, and \cite{hyun2020varsr, SISR9_SRFlow} enables the generation of multiple plausible SR samples.
% where flow-based networks estimate the posterior distribution via invertible blocks \cite{glow, realnvp}.
%

% SISR8_RankSRGAN} and generative prior methods \cite{SISR13_PULSE, SISR10_GLEAN} has arisen, generating photo realistic super-resolution images.
%
%
% Flow based
% In order to tackle the ill-posedness of image SR , normalizing flow-based methods \cite{, }  enabling to sample multiple high-resolution images from the learned distribution. Meanwhile, , and adaptive target generation \cite{jo2021tackling_illposed_1} has improved perceptual quality by generating targets to accept plausible predictions.
%
%
%
% Distillation
%

\section{Limitation}
% V2
It should be noted that Eq.\eqref{eq:final_w_mixup} cannot disentangle the inherent noise term and the estimation error term. Thus, it reintroduces the inherent noise back into the training in later steps. Despite this, experiments emphasize the critical role of stability during the initial steps, setting a strong foundation that leads to overall performance gains. However, we acknowledge the opportunity for further advancement especially for the later training steps, which we leave for future work.

\section{Conclusion}
In this work, we have analyzed the underlying components of vanilla training and systematically developed the current training process.
As a first step, we have disentangled the original loss function into two fundamental components; the centroid and the noise term.
It turns out that the inherent noise term, induced by the ill-posed nature, can potentially raise additional difficulty in vanilla training.
To overcome this issue, we estimate the centroid of all possible high-resolution images and obtain a noise-free lower bound of the original loss function which leads to a well-behaving optimization landscape with enhanced Lipschitzness. 
We further provide an effective method to overcome the limitation of estimation errors, which can be simply adapted into current methods within a few lines of code. 
Experimental results lead us to conclude that the proposed training framework can indeed lead to favorable results.
%
% Based on the analysis of the undiscovered factors, we provide 
% of pixel-wise optimization in image super-resolution and anticipate inspiration for further research.
% We aim to shed light on the underlying components of vanilla training in the context of SISR tasks and 
% As a first step,

\section*{Acknowledgments}
This work was supported in part by MSIT/IITP (No. 2022-0-00680, 2019-0-00421, 2020-0-01821, 2021-0-02068), and MSIT\&KNPA/KIPoT (Police Lab 2.0, No. 210121M06).

\bibliography{reference}

\input{appendix}

\end{document}

%% file: appendix.tex
\appendix
\newpage

\onecolumn
\begin{center}
    \vspace*{0.5cm}
    \LARGE{\bf{Supplementary Material \\ \Large{Noise-free Optimization in Early Training Steps for Image Super-Resolution}}} \\
    \vspace*{0.1cm}
    \bf{\Large{MinKyu Lee, Jae-Pil Heo\footnotemark[1]}} \\
    \vspace*{0.1cm}
    \large{\normalfont{Sungkyunkwan University}} \\
    \large{\normalfont{\{bluelati98, jaepilheo\}@skku.edu}}
    \vspace*{1.0cm}
\end{center}

\footnotetext{$^\ast$Corresponding author}

\begin{figure*}[h]
    \begin{center}
    \includegraphics[width=\textwidth]{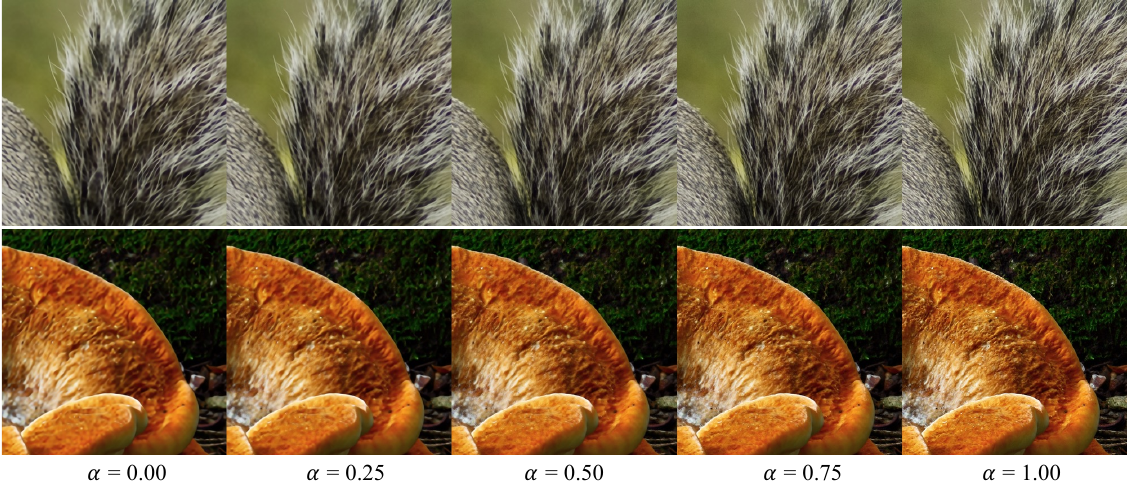}
    \end{center}
    \vspace{-5pt}
    \caption{Visualization of target images in Eq.\eqref{eq:final_w_mixup} as $\alpha$ gradually changes. Unpredictable high-frequency components that can lead to unstable optimization are removed when $\alpha=0$. \textbf{Zoom in for best view.}}
    \label{fig:alpha_visualize}
\end{figure*}

\section{Visual Examples of Target Images\hfill\phantom{PLACEHOLDER}}

In order to perform noise-free training, the proposed training framework utilizes different target images as training proceeds. Specifically, the HR image and the SR image of a pretrained network are blended based on a scheduling parameter $\alpha$. In Fig.\ref{fig:alpha_visualize}, we provide visual examples of the target images in Eq.\eqref{eq:final_w_mixup} as $\alpha$ gradually changes from 0 to 1. As $\alpha$ increases, images become sharper but contain unpredictable high-frequency components, which can potentially lead to noisy and unstable training.

\section{Regressing the Inherent Noise\hfill\phantom{PLACEHOLDER}}
\begin{figure}[h]
    \includegraphics[width=\columnwidth]{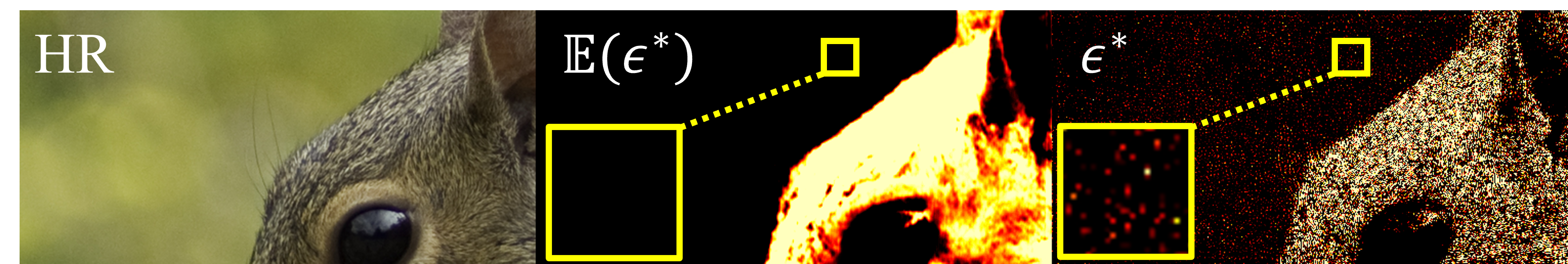}
    \caption{
    Visualization of estimated $\mathbb{E}(\epsilon^*$) and $\epsilon^*$. It can be seen that $\epsilon^*$, which corresponds to the proposed method ECO, can better spot fine-grained noise factors including almost invisible noise in the flat background region. Values are scaled for better visualization. \textbf{Zoom in for best view}.
    }
    \label{fig:noise_comparison}
\end{figure}

To determine the inherent noise $\epsilon^*$ of an HR image, a naive approach might involve training a network to regress the error term. Here we compare this naive approach of regressing the error against the proposed method ECO. The key distinction lies in the way each method approximates the noise.
Notably, the consequence of the regression is approximating the \textit{expectation} of the error $\mathbb{E}(\epsilon^*)$, given an \textbf{LR}. 
In contrast, ECO estimates $\epsilon^*$ directly, by utilizing \textbf{HR} at training time.
In Fig.\ref{fig:noise_comparison}, we visualize estimated $\mathbb{E}(\epsilon^*)$ and $\epsilon^*$. Here, $\mathbb{E}(\epsilon^*)$ is obtained by training an RRDB that is trained to regress $\epsilon^*$ instead of the SR image. It can be observed that $\mathbb{E}(\epsilon^*)$ results in a flat uncertainty map over the entire uncertain region. In contrast, $\epsilon^*$ better spots fine-grained noise factors, including almost invisible noise factors in the flat background. Note that we have shown how this noise can harm the training, underscoring the critical need for precise per-instance estimation of $\epsilon^*$.
\textbf{In a practical view}, estimating $\mathbb{E}(\epsilon^*)$ leads to significantly increased computational cost during training since it requires an additional network, whereas ECO only requires negligible cost. Specifically, the pretrained network $\hat{f}$ can be any off-the-shelf SR network for practical use cases, and $\mu_{emp}$ can be preprocessed before the actual training.

\section{Analysis in the Spectral Domain\hfill\phantom{PLACEHOLDER}} \label{par:gradient_in_spectral_domain}

\begin{figure*}[h]
    \begin{center}
    \includegraphics[width=\textwidth]{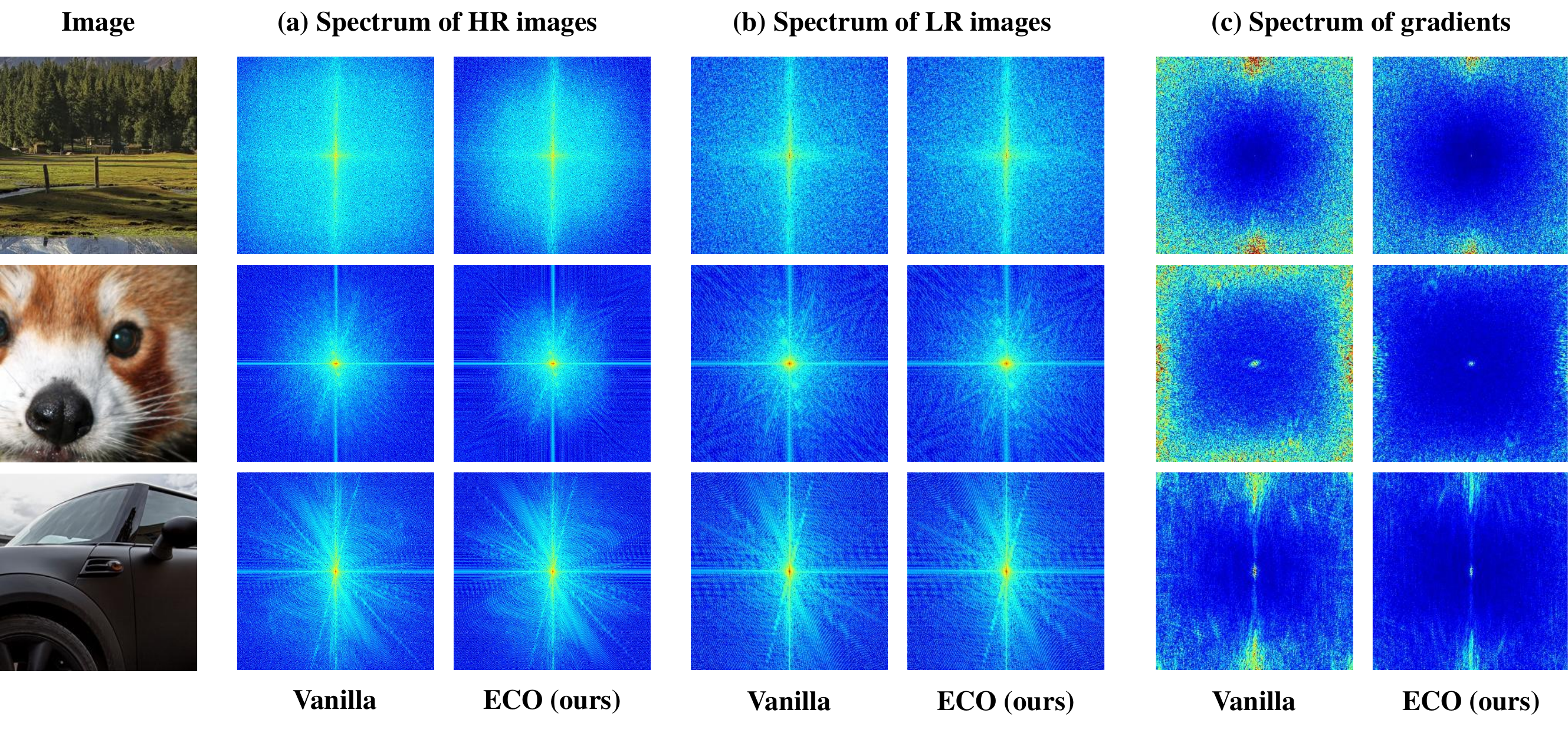}
    \end{center}
    \vspace{-10pt}
    \caption{Spectral analysis of training data pairs and gradients at $\alpha=0$. (a) High-resolution images in the spectral domain. (b) Low-resolution images in the spectral domain. (c) The gradient of the loss in the spectral domain.}
    \label{fig:high_frequency_gradient}
\end{figure*}

We provide further analysis to identify the specific components of image instances that affect the optimization procedure. To achieve this, we applied Fast Fourier Transform (FFT) directly followed by inverse FFT (IFFT) to the images before feeding them into the super-resolution network.
Fig.\ref{fig:high_frequency_gradient}.(a-b) illustrates HR, LR image pairs in the spectral domain and the gradients of losses in the spectral domain, both at $\alpha=0$.
Here, high activation on specific frequency regions indicates that the corresponding components are responsible for the loss values.
In the case of vanilla training, the gradients exhibit strong activation, particularly on the regions where \textit{very} high-frequency components exist. On the other hand, in ECO, while it is well activated on the major frequency components required for recovery, it shows relatively lower activation on very high-frequency components.
We interpret this as indicating the presence of inherent noise terms in the frequency domain.
By this observation, we conclude that ECO, indeed, has a powerful impact on the gradients, especially in regions where inherent noise terms are expected to reside.

\section{Experimental Details\hfill\phantom{PLACEHOLDER}}

\paragraph{Dataset details.}
For all datasets,  we generate LR images with the bicubic function in \text{MATLAB} with the antialiasing option set as true. We have verified that all LR and HR images match the images provided in the prior work \cite{SISR2_EDSR}. In the case of real-world SR, we add zero-mean color Gaussian noise with $\sigma=0.01$ to synthesize real-world images on flight. HR images were cropped modulo the scale factor in order to ensure that HR images match the output size of SR images.

\paragraph{Evaluation details.}
We have compared the proposed training scheme against vanilla training over standard benchmark datasets: Set5 \cite{set5}, Set14 \cite{set14}, BSD100 \cite{bsd100}, Urban100 \cite{urban100} and Manga109 \cite{manga109}. EDSR \cite{SISR2_EDSR}, RCAN \cite{SISR4_RCAN} and SwinIR \cite{liang2021swinir} were used as the representative baseline methods. Performances were evaluated in terms of PSNR and SSIM indices on the Y channel (luminance channel) in the YCbCr space and pixels up to the scale factors in the border were ignored. In the case of real-world SR, we provide average results of 10 different evaluation runs, where the test images were preprocessed for fair comparison.

\paragraph{Implementation details.}
For all experiments, we reproduce representative baseline networks EDSR-baseline, EDSR, RCAN and SwinIR with both vanilla training and our method. 
To train the networks, we use 800 RGB images from DIV2K \cite{div2k} and images were preprocessed as sub-patches for faster I/O. Note that we have only used the DIV2K dataset (instead of the DF2K) also for SwinIR, and do not use exponential moving averaged weights for both the baseline method and our method.
The patch size of low-resolution images was kept as 48$\times$48 for all scale factors as in prior works. Random horizontal and vertical flips were used, together with 90$^\circ$, 180$^\circ$, 270$^\circ$ random rotations as basic training augmentation. 
%
%
% ------------ training details
In Table.(1), we follow \cite{rcanit} which demonstrated comparable performance while significantly reducing the required training time.  We increase the learning rate by $\times$16 and the mini-batch size by $\times$8, decreasing the total training iteration by $\times$8. Specifically, the mini-batch size is 128, the learning rate is 0.0016, and the total training iteration is 125K.
We also substitute the scheduler as cosine annealing \cite{cosineannealing} and utilize the Lamb \cite{lamb} optimizer which is known to work better on larger batch sizes. We train our networks on two NVIDIA TITAN RTX GPUs and the batch size was selected to fit GPU memory. We train networks from scratch for $\times$2 SR. For $\times$3 SR and $\times$4 SR, we start from the pretrained weight of the $\times$2 SR network. 
However, for $\times$4 RCAN (both vanilla training and ours), we kept the setting of the original works \cite{SISR4_RCAN} since the baseline models produced undefined numbers (NaN) with larger learning rates. Specifically, the mini-batch size is 16, the learning rate is $10^{-4}$, the total training iteration is 1000K and the Adam optimizer was used and the learning rate was reduced by half every 200K iterations.
All networks in Tab.(1) were trained with two NVIDIA A6000 and all other networks were trained with one NVIDIA TITAN RTX.